\documentclass{INTERSPEECH2024}

\interspeechcameraready 

\usepackage{array}
\usepackage{tabularx}
\usepackage{tabulary}
\usepackage{makecell}

\usepackage{multirow}
\usepackage[skip=1pt]{caption}
\usepackage{booktabs}
\usepackage{threeparttable}

\usepackage[dvipsnames]{xcolor}

\usepackage[natbib, bibencoding=utf8, citestyle=numeric, bibstyle=ieee, maxbibnames=999, maxcitenames=2, mincitenames=1, sortcites]{biblatex}
\bibliography{bibliography}

\usepackage{wasysym}

\usepackage[activate]{microtype}
\newcommand{\eg}{e.\,g.}

\usepackage[acronym, shortcuts, nohypertypes={acronym}]{glossaries}
\newacronym{AI}{AI}{artificial intelligence}
\newacronym{ASR}{ASR}{automatic speech recognition}
\newacronym{compare}{{ComParE}}{Interspeech Computational Paralinguistics ChallengE feature set}
\newacronym{COPD}{COPD}{chronic obstructive pulmonary disease}
\newacronym{egemaps}{{eGeMAPS}}{extended Geneva minimalistic acoustic parameter set}
\newacronym{FNR}{FNR}{False Negative Rate}
\newacronym{FPR}{FPR}{False Positive Rate}
\newacronym{MAUS}{MAUS}{Munich AUtomatic \textbf{Segmentation}}
\newacronym{ML}{ML}{machine learning}
\newacronym{NuS}{NuS}{the North Wind and the Sun [\textit{Der Nordwind und die Sonne}]}
\newacronym{RBF}{RBF}{radial basis function}
\newacronym{SHAP}{SHAP}{SHapley Additive exPlanations}
\newacronym{SVM}{SVM}{support vector machine}
\newacronym{UAR}{UAR}{unweighted average recall}

\usepackage[capitalise, nameinlink, noabbrev]{cleveref}

\title{
Sustained Vowels for Pre- vs Post-Treatment COPD Classification
}

\name[affiliation={1}]{Andreas}{Triantafyllopoulos}
\name[affiliation={1}]{Anton}{Batliner}
\name[affiliation={2}]{Wolfgang}{Mayr}
\name[affiliation={2}]{Markus}{Fendler}
\name[affiliation={1,3}]{Florian}{Pokorny}
\name[affiliation={4}]{Maurice}{Gerczuk}
\name[affiliation={1}]{Shahin}{Amiriparian}
\name[affiliation={2}]{Thomas}{Berghaus}
\name[affiliation={1,4,5}]{Björn}{Schuller}
\address{
  $^1$CHI -- Chair of Health Informatics, MRI, Technical University of Munich, Germany \\
  $^2$Department of Cardiology, University Hospital Augsburg, University of Augsburg, Germany \\
  $^3$Division of Phoniatrics, Medical University of Graz, Austria \\
  $^4$Chair of Embedded Intelligence for Health Care and Wellbeing, University of Augsburg, Germany \\
  $^5$GLAM -- Group on Language, Audio, \& Music, Imperial College, UK
  }
\email{andreas.triantafyllopoulos@tum.de}

\email{andreas.triantafyllopoulos@tum.de}

\keywords{COPD, computational paralinguistics, sustained vowels, digital health, treatment evaluation}

\begin{document}

\maketitle

\begin{abstract}
Chronic obstructive pulmonary disease (COPD) is a serious 
inflammatory lung disease affecting millions of people around the world.
Due to an obstructed airflow from the lungs, it also becomes manifest in patients' vocal behaviour.
Of particular importance is the detection of an exacerbation episode, which marks an acute phase  and often requires hospitalisation and treatment.
Previous work has shown that it is possible to distinguish between a pre- and a post-treatment state using automatic analysis of read speech.
In this contribution, we examine whether sustained vowels
can provide a complementary lens for telling apart these two states.
Using a cohort of 50 patients, we show that the inclusion of sustained vowels can improve performance to up to 79\% unweighted average recall, from a 71\% baseline using read speech.
We further identify and interpret the most important acoustic features that characterise the manifestation of COPD in sustained vowels.
\end{abstract}

\glsresetall

\section{Introduction}
\label{sec:intro}

Chronic obstructive pulmonary disease (COPD) -- mainly caused by smoking \citep{Yoshida07-POC} -- is the fifth leading cause of death worldwide \citep{WHO20} and represents a major socioeconomic burden.
Its rapid and accurate monitoring using artificial intelligence (AI) has great potential in assisting medical practitioners and facilitating better treatment outcomes.
As a respiratory disease, it substantially affects vocalisations, making it a prime target for speech-based analysis.
An exacerbation phase often requires hospitalisation \citep{Singh19-GSF,Walters14-SCF}.
The definition of such a 
phase and a subsequent evaluation of treatment heavily rely on clinical assessments, exacting a huge toll on medical resources and being 
rather subjective.
An automated tool to detect whether patients are still in an exacerbation phase or ready to be released could vastly improve quality of service and reduce the strain on medical practitioners.

To that end, automatic voice and speech analysis in respiratory diseases has gained momentum with the COVID-19 pandemic.
A still open question is which vocal phenomena are most promising for employing them as proxy: When we aim at an unintrusive recording scenario, we can for instance choose amongst voice and speech phenomena such as breathing, coughing, \emph{isolated/sustained vowels} \citep{bartl2021voice},   
or \emph{connected speech} (read or spontaneous). 
Read or free speech for distinguishing between COPD patients and healthy individuals as well as between different states of COPD have already been addressed in previous work~\citep{Mohamed14-VCI, Merkus20-DEA, Cleres21-LAV, Triantafyllopoulos22-DBP}. 
\citet{Merkus20-DEA} also employed productions of /a:/;  yet, sustained vowels are generally underrepresented in recent studies.
Here, we therefore focus primarily on them and test their complementarity to connected speech.

The research on sustained vowels dates back at least to Siegenthaler in 1950 \citep{Siegenthaler50-ASO}, who calls their three phases
``initiation,  middle  period, and  conclusion''. In the present paper, we use the terms \emph{onset}, \emph{centre}, and \emph{offset} -- `centre' instead of `steady-state' because it is a neutral term and does not imply any specific acoustic profile.
%
Already in 1990, Klingholtz \citep{Klingholtz90-ARO} reported over 300 studies on voice pathologies, most of them using sustained vowels. 
We can speculate about this bias towards sustained vowels: In `pre-computer', traditional phonetics, it surely has been easier to target isolated vowels than connected speech; speech pathology research is based on the same premises \citep{Schuller14-CPE}. In contrast, automatic speech processing soon went from isolated words to connected speech, which remained a natural object of investigation. 
As stated in \citep{Fourcin02-MVI}: 
``Clinical voice measurement [...] is mostly directed towards the appraisal of the ability to produce a sustained vowel.'' This goes together with the practicability of \emph{maximum phonation time} (MPT) as ``noninvasive, fast, and low-budget measurement [...] considered an objective measure of the efficiency of the respiratory mechanism during phonation.'' \citep{Speyer08-MPT}.
As `typical' (cardinal) vowels, the most frequent candidate seems to be /a:/, followed by /i:/ and/or /u:/, and sometimes alternatively or additionally /e:/ and/or /o:/. 
If MPT is not targeted, to avoid inconsistencies due to phonation onset and offset, often the middle part of the vowel is employed.
Furthermore, \citet{Amir09-AAO}   
found only minimal differences between a steady state phase and the entire vowels /a:/ and /i:/ when modelling pathological subgroups vs healthy speakers. Note that onset and offset were defined based on the shape of the intensity contour; thus, they  were  mismatched for the different vowels and did not disentangle the role of MPT.


In recent times, connected speech is gaining traction in automatic analysis.
In \citep{Parsa01-ADO},
classification results ``were comparable for measures extracted from continuous speech samples and for those based on sustained vowels''. 
\citet{Maryn10-TIE}
argue that continuous speech and sustained vowels should be combined in the analysis of disordered voice, 
whereas \citet{Moon12-MOA}
conclude that sustained vowels cannot be a substitute for `real-time' phonation; see as well \citep{Cordeiro15-CSC,Wang22-CSF}.
%

In the present study, we aim to fill research gaps on the basis of a COPD pre- vs post-treatment design.
Previous studies have not addressed all cardinal vowels and all different segmentations (whole vowel, cut-outs such as onset, centre, offset) at the same time.
By jointly accounting for multiple cardinal vowels, and systematically varying their temporal segmentation, we attempt to disentangle the role of MPT and acoustic parameters on the manifestation of COPD in sustained vowels.
We further contrast that with COPD's impact on connected speech in an attempt to compare the two data collection protocols.
%
Another aim of our study is to find out more about the acoustic feature characteristics of pre- and post-treatment COPD -- which, in turn, might be representative for the manifestation of respiratory diseases in recorded speech signals in general.

\section{Methodology}
\label{sec:methodology}

\noindent
\textbf{Data:} Our dataset includes a total of 50 (male: 26; female: 24) COPD patients recorded at the University Hospital Augsburg.
The patients produced 
the five German cardinal vowels (/a:/, /e:/, /i:/, /o:/, /u:/; in that order), followed by reading out loud Aesop's the ``North Wind and the Sun'' (NaS) in German --  both before (\emph{pre-treatment}) and after (\emph{post-treatment}) treatment.
%
At both times, the patients answered a standardised patient questionnaire with the modified BORG scale (assessing the degree of dyspnoea) \citep{Kendrick00-UOT} and the CAT scale (measuring health-related quality of life) \citep{Gupta14-TCA} and underwent pulmonary function testing (PFT) before discharge.
All of them had a diagnosed COPD with a history of smoking.
The mean duration of the COPD disease was $9.64$ Years (N: $42$) with a minimum duration of $3$ and an maximum duration of $22$ years.
According to the new COPD-ABE classification, all included participants were categorised into patients group E~\citep{Agusti23-GIF}.
The mean exacerbation rate in the last two years before recruitment was $2.3$ ($\pm$ $3.4$) with a maximum of $17$ exacerbations and a minimum of $0$.

All of the patients were suffering of an acute exacerbation at the time of inclusion. 
$98\%$ received treatment with steroids, $58\%$  with additional antibiotics.
Pre- and/or inner clinical ventilation therapy was performed at $54\%$ of the patients, $32\%$ were already provided with ventilation at home and $76\%$ with long term oxygen therapy (LTOT). 
The mean time interval between the 1\textsuperscript{st} and the 2\textsuperscript{nd} recording was $9.1$ days ($\pm 4.31$).
The mean CAT Score at recruitment was $28$ ($\pm 5.44$) and the perceived dyspnoea after the BORG scale was $4.40$ ($\pm 2.78$).
At the time of discharge, both scales showed an improvement to $22.38$ ($\pm 5.97$) for the CAT Score and to $2.10$ ($\pm 1.79$) for the BORG score.


Patients were instructed to sustain the vowels as long as possible, by that indicating MPT; 
the vowels had an average duration of $4.71 \pm 3.78$s pre- and $6.52 \pm 4.17$s post-treatment.
Each recording contains all vowels, which are segmented manually by the 2\textsuperscript{nd} author using PRAAT \citep{Boersma22-PDP}.
Additionally, we segmented NaS by first using the \ac{MAUS} system to align the audio of each patient with the (target) transcript~\citep{schiel1999,kisler2017multilingual} and subsequently split the story into 20 \textbf{prosodic phrases} (\emph{phrase-level}), 
as has been done in~\citep{Triantafyllopoulos22-DBP}.

\noindent
\textbf{Features:} We use openSMILE~\citep{eyben2010opensmile} to extract the extended Geneva Minimalistic Acoustic Parameter Set (eGeMAPS)~\citep{eyben2015geneva} as a small-scale  feature set (N: 88) previously shown useful to characterise COVID-19 on the basis of vowels~\citep{bartl2021voice} and COPD on the basis of read speech~\citep{Triantafyllopoulos22-DBP}. This allows us to use a consistent feature set across different vocalisations.
It computes a set of functionals (e.\,g., mean, coefficient of variation, minimum, maximum, etc.) over a number of low-level descriptors (LLDs): F0, H1-H2 (amplitude difference between first and second F0 harmonics, relative to F0 amplitude), formant frequencies (F1, F2, F3), H1-A3 (amplitude difference between first F0 harmonic and maximum harmonic in F3 range, relative to F0 amplitude), harmonic-to-noise ratio (HNR), jitter, shimmer, loudness, spectral slope, Hammarberg index, and alpha ratio.
These features are  extracted i) for the entire vowel (\emph{Whole}), ii) for the different segmentations outlined below, and iii) for the prosodic phrases of NaS.

\noindent
\textbf{Mismatched segmentation:}
Our first segmentation allows for mismatched vowel durations, thus implicitly accounting for post-treatment speakers holding their vowels longer.
We split the signals in three parts, computed separately for each vowel:
onset (first 33\%), centre (middle 33\%), and offset (last 33\%).

\noindent
\textbf{Matched segmentation:}
To alleviate the impact of mismatched durations, we extract matched segments where we keep a constant time window of $w=\{1, 2, 3\}$ seconds.
Assuming a signal of length $N$, this window is once again extracted from the onset ($[0:0+w]$), the centre ($[N/2-w/2:N/2+w/2]$), and the offset ($[N-w:N]$).
Signals with $N<w$ are mirrored to match the required duration.
This gives us sequences of matched length where the effect of holding a vowel longer or shorter is removed.
As results were always best when using $w=3$, for brevity, we only report results for this value.

\noindent
\textbf{Modelling:} As our patient cohort is relatively small, we always use a nested leave-one-speaker-out cross-validation, whereby data from every speaker is used exactly once for testing, each time using the data from all other speakers for training.
This process is used to train a \ac{SVM} classifier, where we optimise the cost parameter (\{$.0001$, $.0005$, $.001$, $.005$, $.01$, $.05$, $.1$, $.5$, $1$\}) and kernel function (\{linear, polynomial, \ac{RBF}\}) in a grid search manner.
These parameters are always optimised on the development partition, which is created by splitting the training speakers into two speaker-disjoint sets.
Once the optimal set of parameters is identified (based on development set performance), we train a final model on the entire training data for each fold, for which we report performance.

\noindent
\textbf{Normalisation:} We contrast standard normalisation, where parameters are computed on the entire training set and applied to the validation/test set, with speaker normalisation, where parameters are computed separately for each speaker.
This form of normalisation can remove individual effect, and was found to make a significant performance difference in previous COPD-related work~\citep{Triantafyllopoulos22-DBP}.


\noindent
\textbf{Evaluation: }
Performance is evaluated using the unweighted average recall (UAR), the added recall of all classes divided by their number.
In the case of vowels, preliminary results showed that training on each vowel independently leads to inconsistent results, often near chance-level (50\%).
Therefore, we always train on \emph{all five vowels} and aggregate (max-vote) the predictions for each session.
Similarly, we train on all 20 phrases and aggregate the predictions for them for each story.
Thus, results for the two types of speech material are directly comparable, with one prediction for each session (pre- vs post-treatment) per speaker, resulting in a \textbf{SE}ssion-level UAR (SE\textsubscript{UAR}).
%
Additionally, we provide a disaggregated evaluation for males (\mars) and females (\female) to better understand how our models behave across gender.
Finally, an alternative lens to understand model behaviour is the analysis of \textbf{SP}eaker-level performance (SP\textsubscript{UAR})~\citep{Triantafyllopoulos22-DBP}; there, we compute the UAR by only considering the instances belonging to each single speaker in isolation, separately for the phrases and the vowels.


\section{Classification Performance}
\label{sec:results}

\begin{table}[t]
    \centering
    \caption{
    Leave-one-speaker-out UAR [\%] when using standard and speaker normalisation for phrases, mismatched/matched segmentation, and a late fusion (max-vote) of results.
    CIs in squared brackets.
    Additionally showing performance stratified per males (\mars) and females (\female). 
    Chance level 50\%.
    }
    \label{tab:results}
    \resizebox{\columnwidth}{!}{
    \begin{tabular}{l|c|cc}
        \toprule
        \textbf{Norm.} & \textbf{Standard} & \multicolumn{2}{c}{\textbf{Speaker}}\\
        \textbf{Unit} & \textbf{SE\textsubscript{UAR}} & \textbf{SE\textsubscript{UAR}} & \textbf{\mars\textsubscript{UAR}/\female\textsubscript{UAR}}\\
        \midrule
        \textbf{Phrases (A)} & 56 [49-67] & 71 [61-80] & 62/81\\
        \midrule
        \multicolumn{4}{c}{\textbf{Mismatched segmentation}}\\
        \midrule
        \textbf{Onset}    & 56 [47-65] & 75 [67-83] & 75/75 \\
        \textbf{Centre}   & 62 [52-72] & 70 [61-78] & 71/69 \\
        \textbf{Offset}   & 63 [54-73] & 73 [64-82] & 79/67 \\
        \textbf{Whole}    & 64 [55-74] & 72 [63-80] & 75/69 \\
        \textbf{MPT (B)}  & 58 [48-69] & 69 [59-78] & 63/75 \\
        \midrule
        \multicolumn{4}{c}{\textbf{Matched segmentation}}\\
        \midrule
        \textbf{Onset}       & 58 [49-68] & 69 [59-77] & 77/60\\
        \textbf{Centre}      & 58 [49-68] & 61 [52-70] & 69/52\\
        \textbf{Offset (C)}  & 54 [45-64] & 70 [61-78] & 77/62\\
        \midrule
        \multicolumn{4}{c}{\textbf{Late Fusion}}\\
        \midrule
        A + B     & 56 [47-65] & 71 [63-79] & 63/79\\
        A + C     & 57 [49-66] & 70 [62-78] & 65/75\\
        B + C     & 55 [46-65] & 72 [64-80] & 71/73\\
        A + B + C & 54 [44-65] & \colorbox{lightgray}{\textbf{79 [71-86]}} & 73/85\\
        \bottomrule
    \end{tabular}
    }
\end{table}

\begin{figure}[t]
    \centering
    \includegraphics[width=\columnwidth]{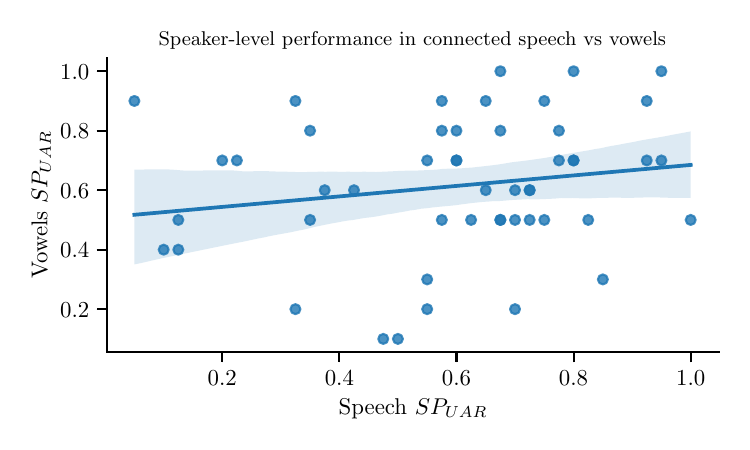}
    \caption{
    Comparing $SP_{UAR}$ for vowels vs connected speech.
    $SP_{UAR}$ for connected speech computes UAR over all phrases of each speaker~\citep{Triantafyllopoulos22-DBP}.
    $SP_{UAR}$ for vowels computes UAR over all 5 vowels of each speaker; using 3-sec window at onset.
    Spearman's $\rho$ for the two models: $.2$ (p-value: $.16$).
    }
    \label{fig:speakerlevel}
\end{figure}

Our results are presented in \cref{tab:results},
together with 95\% confidence intervals (CIs) computed over 1000 bootstrap samples.
We first observe that speaker normalisation leads to considerable gains across all combinations, similarly  to~\citep{Triantafyllopoulos22-DBP}; we therefore focus our discussion on those.
Phrase-level performance is at 71\%, lower than the 80\% in \citep{Triantafyllopoulos22-DBP}, but we are 
dealing with a larger -- and possibly more diverse -- patient cohort.

Vowel results obtained with a mismatched segmentation are comparable across the different parts, ranging from 75\% when using the onset to 70\% when using the centre.
Remarkably, MPT alone is sufficient for acquiring a UAR of 69\%, indicating that this feature is a powerful predictor of COPD.
Enforcing a matched segmentation lowers performance, now ranging from 70\% in the case of offset to 61\% for centre; the latter drop being a by-product of removing the effect of MPT.
However, onset and offset are still competitive and outperform the centre, potentially benefiting from transition effects.
This is in contrast to previous work~\citep{Amir09-AAO} where no additional gains were reported by including transitions in the modelling of /a:/ and /i:/. 

The complementarity of acoustic predictors of COPD in sustained vowels vs read speech and MPT is tested by a late fusion of their results.
We pick offset as the best-performing segmentation in matched durations, as this disentangles the impact of MPT.
The fusion of phrase-level, vowel, and MPT results obtains a best performance of 79\% UAR.
Further insights in this aspect of complementarity can be found in \cref{fig:speakerlevel}, which shows the SP\textsubscript{UAR} obtained via vowels vs that obtained using connected speech.
Although there is some agreement between the two classifiers (Spearman's $\rho$: .2), there are several speakers for which one model performs well and the other misclassifies.
This indicates that the two types of vocalisations can complement one another, and that the use of vowels can provide additional benefits to employing read speech.

Interestingly, there is a clear difference between the performances in {\female} vs \mars, especially for phrases vs matched segmentation: In \mars, performance for the reading task is clearly worse with 62\% UAR [48\%-75\%] as compared to 81\% UAR [69\%-92\%] in \female, whereas for matched segmentation it is the other way round, with offset yielding 77\% UAR [66\% - 88\%] and 62\% UAR [49\% - 76\%] in {\mars} and \female, respectively.
Now, {\female} showed a better response to treatment, as seen in the difference between CAT/BORG scores: 
The mean difference for CAT was 5.3 (\mars: 5.1; \female: 5.8) and for BORG 2.0 (\mars: 1.9; \female: 2.2).
This is even though {\female} were hospitalised on average for a (slightly) shorter time (\mars: 7.7 days; \female: 7.4 days). 
This could explain the difference in the reading task but not the one for matched segmentation. 
We might further speculate that {\female} are more `literate' than {\mars} and, thus, better at the reading task when recovered from exacerbation,  showing a bigger difference to their pre-treatment state;
some evidence for women being more `canonical' than men can be found in \citep{Trudgill72-SCP,Kreiman11-FOV,Hoenig14-AMM}. 

\section{Feature Interpretation}

\begin{figure}[!htbp]
    \centering
    \includegraphics[width=\columnwidth]{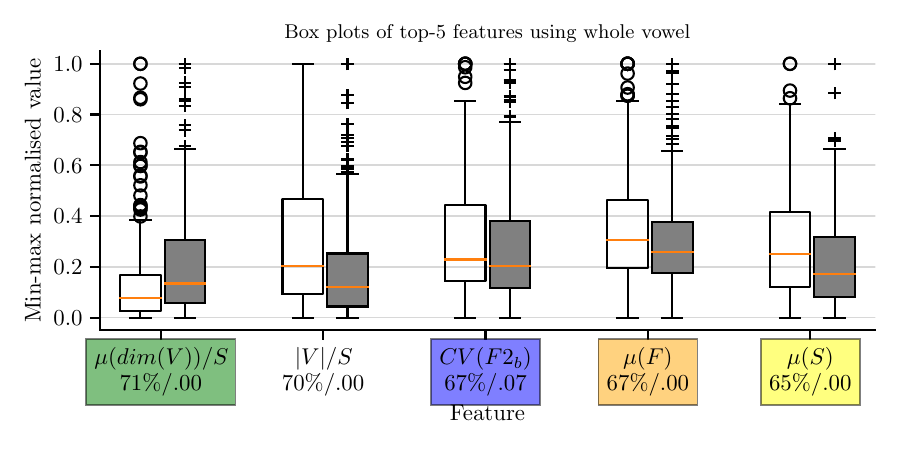}
    \includegraphics[width=\columnwidth]{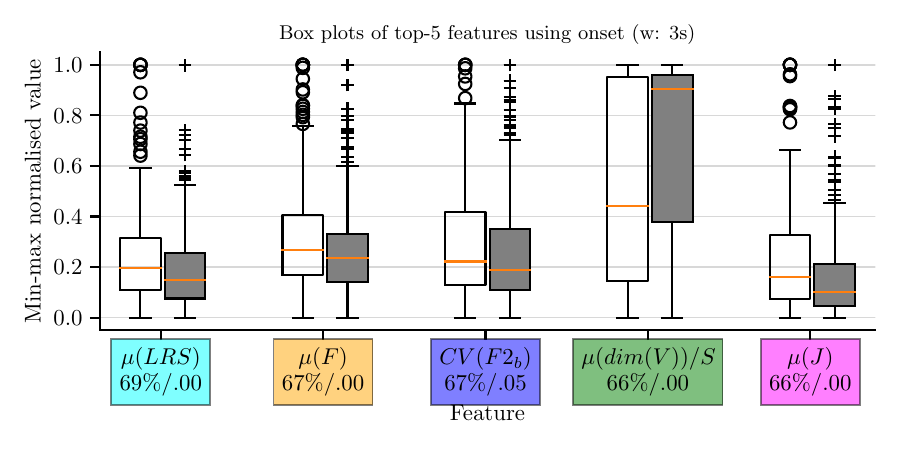}
    \includegraphics[width=\columnwidth]{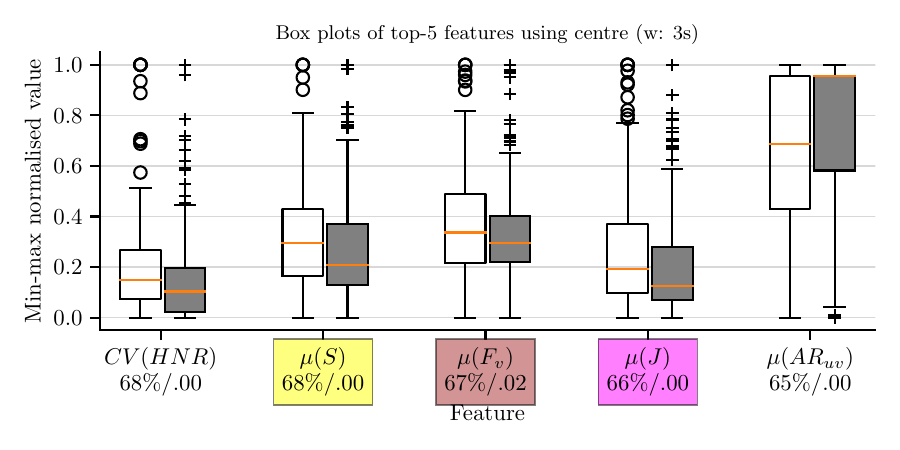}
    \includegraphics[width=\columnwidth]{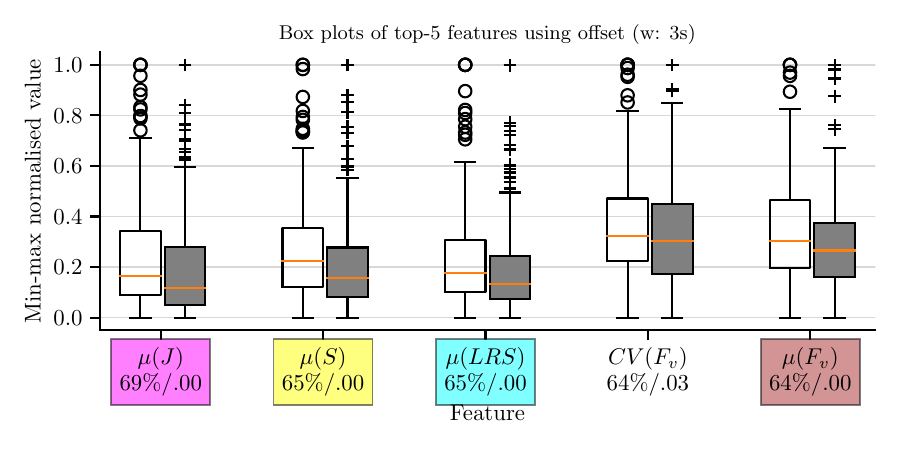}
    \caption{
    Top-5 features 
    identified by training 
    them individually 
    using the entire segments of all vowels and speaker normalisation,
    extracted pre- (solid; left) and post-treatment (dashed; right).
    Showing UAR [\%] and p-values from two-sided Mann-Whitney U test.
    $\mu(dim(V))$: mean voiced segment length [in seconds];
    $|V|/S$: number of voiced segments per second;
    $CV(F2_b)$: coefficient of variation of F2 bandwidth;
    $\mu(F)$: mean spectral flux;
    $\mu(S)$: mean shimmer;
    $\mu(LRS)$: mean loudness rising slope;
    $\mu(J)$: mean jitter;
    $CV(HNR)$: coefficient of variation of harmonic-to-noise ratio;
    $\mu(F_v)$: mean spectral flux in voiced regions;
    $\mu(AR_{uv})$: mean alpha ratio in unvoiced regions.
    Features that appear multiple times are coloured.
    }
    \label{fig:boxplot}
\end{figure}

To further understand how COPD manifests in sustained vowels, we interpret the 5 most important features for different segmentations. 
These are identified by training a model on each individual eGeMAPS feature using the exact same configuration as for the model trained on all features.
We perform this for the entire vowels, as well as for the onset, centre, and offset in the case of matched segmentation.
We visualise them for the two states in \cref{fig:boxplot} and compute two-sided Mann-Whitney U test \textit{p}-values (without correction for repeated measurements).
Note that we do not advocate using \textit{p}-values for evaluating results, due to the inherent problems of Null Significance Testing~\citep{Wasserstein16-TAS}.

Showing the top-5 features when using the entire vowels reveals the implicit role of MPT.
The top two features are the \textbf{mean voiced segment length ($\mu(dim(V))$)} and the \textbf{number of voiced segments per second ($|V|/S$)}, same as found for the differentiation of individuals with and without COVID-19~\citep{bartl2021voice}.
$\mu(dim(V))$ is higher post-treatment, whereas $|V|/S$ is lower.
This is a by-product of 
phonation duration, as longer phonation post-treatment resulted in higher 
 $\mu(dim(V))$, even though the total number of voiced segments was less (more regular phonation in post-treatment results in less breaks in F0).
This is also evident in their correlation with MPT, which is $.51$ and $-.33$ for $\mu(dim(V))$ and $|V|/S$, respectively.
Matched segmentation alleviates its impact; $\mu(dim(V))$ only arises once (for onset), but with a lower performance and more dispersion for pre- and post-treatment.

Overall, several features appear multiple times:
\textbf{Shimmer} (S) and \textbf{Jitter} (J) appear thrice each, and are always lower post-treatment, similar as for connected speech~\citep{Triantafyllopoulos22-DBP}, 
indicating a  more stable phonation, with less period-to-period fluctuations.  
\textbf{Spectral flux} (F) appears in all constellations, either computed over all segments or over voiced segments only, and is also always lower post-treatment.
Flux measures spectral magnitude changes between two successive frames; the fact that its mean and its coefficient of variation are lower after treatment can be interpreted as speakers keeping a more homogeneous timbre over time.

\textbf{F2 bandwidth} ($CV(F2_b)$) fluctuates less post- vs pre-treatment.
This is consistent with previous work where it has been shown that dysphonic speakers display a broader formant bandwidth \citep{Ishikawa20-BAA}, meaning higher formant dispersion and mutual masking of neighbouring formants and by that vowels~\citep{Cheveigne99-FBA}, a finding also seen in connected speech~\citep{Triantafyllopoulos22-DBP}.

There are also features which appear uniquely for different segmentations.
\textbf{Harmonic-to-noise ratio} (HNR) and \textbf{alpha ratio} (AR) only appear at the central part of the vowels.
HNR variation is higher pre-treatment, which translates to more fluctuations in the energies of signal harmonics over noise harmonics, pointing to harsher voice quality.
AR additionally shows a small shift towards energy in higher frequencies (1\,kHz--5\,kHz over 50\,Hz--1\,kHz) post-treatment.

Finally, the \textbf{mean loudness rising slope} ($\mu(LRS)$) is higher pre-treatment for both the onset and the offset, and is the most important feature for the former.
This captures transient effects in these two parts of the signal, as the feature measures the average (local) minimum-to-maximum slope of the loudness curve, and shows that speakers can better maintain a stable intensity post-treatment.

Overall, the acoustic correlates of COPD in sustained vowels are similar to those found 
for connected speech~\citep{Triantafyllopoulos22-DBP}.
Restricted airflow pre-treatment negatively impacts loudness in both cases, with some temporal effects in vowels relating to onset and offset.
Formants show a higher dispersion, which in connected speech further appears as more imprecise articulation. 
Jitter and shimmer are increased.
Interestingly, pitch did not arise as one of the top five features, while being the most important one  for connected speech~\citep{Triantafyllopoulos22-DBP}.
We might speculate that (short) irregular phonation -- causing F0 extraction errors such as octave jumps or wrong voiced-unvoiced decisions, by that yielding higher pitch values in the pre-treatment condition~\citep{Triantafyllopoulos22-DBP} -- more often occurs in transition phases. 
These are ubiquitous in connected speech but not characteristic for sustained vowels. 
Nevertheless, pitch-based functionals (80\% percentile, coefficient of variation) showed good performance with 64\% for the whole vowel, onset, and centre -- just not as good as the fifth best feature.

\section{Conclusion}
We examined the use of sustained vowels for the classification of post- vs pre-treatment COPD patients and contrasted it with connected speech.
Our \textbf{findings} show that sustained vowels offer complementary benefits to the latter.
We further disentangled the effect of maximum phonation time (MPT) by investigating both matched (same duration) and mismatched (different duration) segmentations of different vowel stages (onset, centre, offset), and show that signal acoustics can provide additional information.
A subsequent interpretation of most important features shows voice quality and intensity measures to be positively affected by treatment, with evidence of temporal effects in the onset and offset phases.
\textbf{Limitations} of our work are the relatively small number of patients -- which is, however, characteristic for speech pathology.
\textbf{Future work} can pursue a tighter integration of different signals ({\eg} speech, vowels, coughing) for a more holistic characterisation of COPD.




\newpage
\section{Acknowledgements}
This work has received funding from the DFG's Reinhart Koselleck project No.\ 442218748 (AUDI0NOMOUS).

\section{\refname}
\printbibliography[heading=none]

\end{document}